\documentclass[10pt,twocolumn,letterpaper]{article}

\usepackage{cvpr}              

\usepackage{graphicx}
\usepackage{amsmath}
\usepackage{amssymb}
\usepackage{booktabs}
\usepackage{tabularx}
\usepackage{bm}
\usepackage{cite}
\usepackage{float}
\usepackage{wrapfig}
\usepackage{calc}
\usepackage[table]{xcolor}
\usepackage{multirow}
\usepackage{tikz}
\usetikzlibrary{positioning}
\definecolor{testcolor}{rgb}{1, 1, 0.6}
\definecolor{myred}{HTML}{B85450}
\definecolor{myblue}{HTML}{6C8EBF}
\definecolor{polaris}{RGB}{121,150,196}
\definecolor{composite}{RGB}{230,163,126}
\definecolor{flickr8k-cf}{RGB}{129,190,142}
\definecolor{flickr8k-ex}{RGB}{211,123,126}
\definecolor{TitleColor}{gray}{0.95}
\definecolor{LightCyan}{rgb}{0.88,0.95,1}

\usepackage[pagebackref,breaklinks,colorlinks]{hyperref}

\usepackage[capitalize]{cleveref}
\crefname{section}{Sec.}{Secs.}
\Crefname{section}{Section}{Sections}
\Crefname{table}{Table}{Tables}
\crefname{table}{Tab.}{Tabs.}

\def\paperID{5033} 
\def\confName{CVPR}
\def\confYear{2024}

\begin{document}

\title{\hspace{-1mm}Polos: Multimodal Metric Learning from Human Feedback for Image Captioning}

\author{
  \begin{tabular}{cccc}
    Yuiga Wada \hspace{7mm} & Kanta Kaneda \hspace{7mm} & Daichi Saito \hspace{7mm} & Komei Sugiura \\
  \end{tabular} \\
  Keio University \\
  \url{https://yuiga.dev/polos}
}

\maketitle
\vspace{-3mm}
\begin{abstract}
\vspace{-4mm}

Establishing an automatic evaluation metric that closely aligns with human judgments is essential for effectively developing image captioning models. Recent data-driven metrics have demonstrated a stronger correlation with human judgments than classic metrics such as CIDEr; however they lack sufficient capabilities to handle hallucinations and generalize across diverse images and texts partially because they compute scalar similarities merely using embeddings learned from tasks unrelated to image captioning evaluation.
In this study, we propose Polos, a supervised automatic evaluation metric for image captioning models.
Polos computes scores from multimodal inputs, using a parallel feature extraction mechanism that leverages embeddings trained through large-scale contrastive learning.
To train Polos, we introduce Multimodal Metric Learning from Human Feedback (M$^2$LHF), a framework for developing metrics based on human feedback. We constructed the Polaris dataset, which comprises 131K human judgments from 550 evaluators, which is approximately ten times larger than standard datasets. Our approach achieved state-of-the-art performance on Composite, Flickr8K-Expert, Flickr8K-CF, PASCAL-50S, FOIL, and the Polaris dataset, thereby demonstrating its effectiveness and robustness.

\end{abstract}
\vspace{-3mm}

\vspace{-3mm}
\section{Introduction}
\vspace{-2mm}

\begin{figure}[t]
    \centering
    \includegraphics[width=\linewidth]{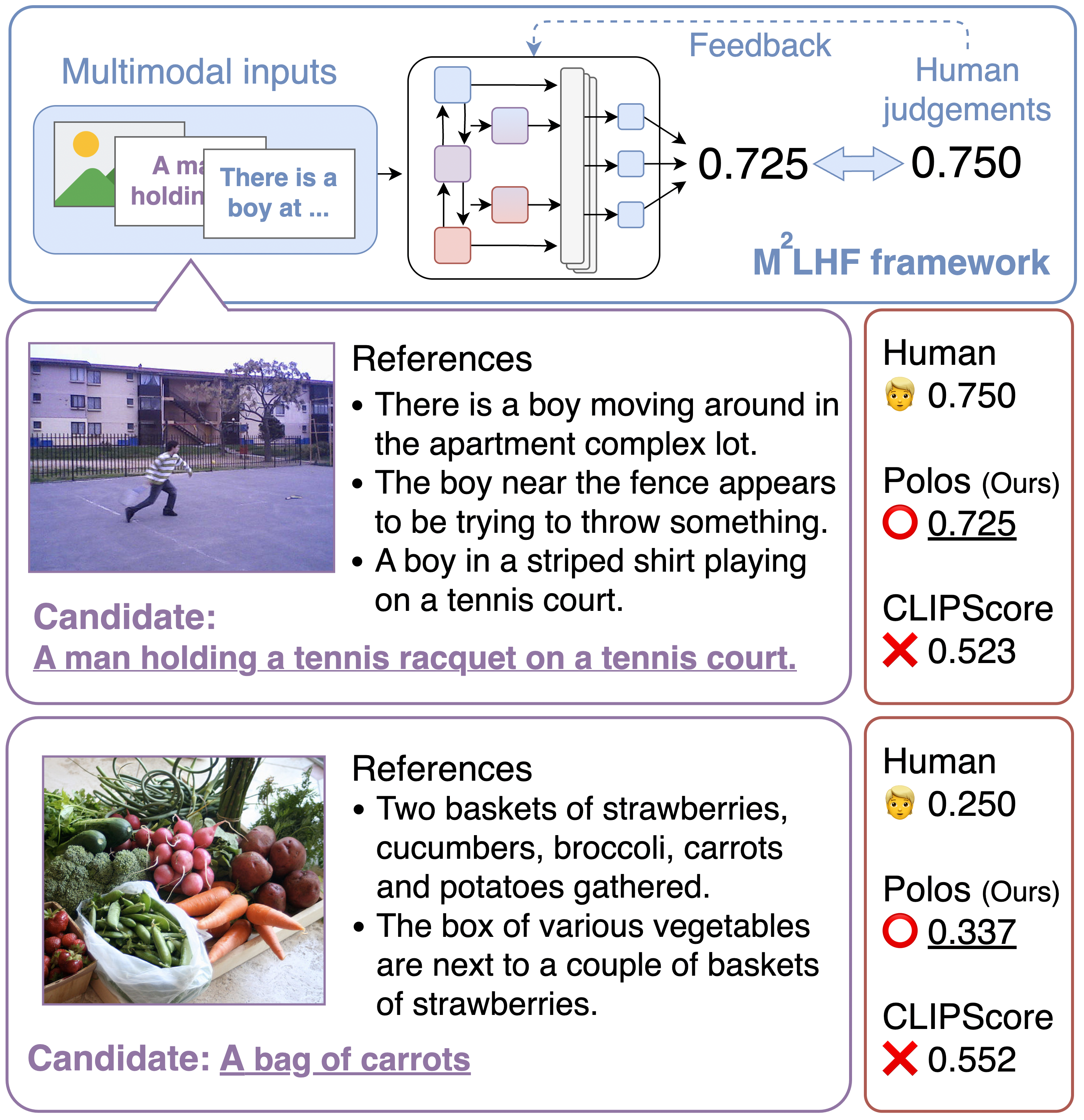}
    \vspace{-5mm}
    \caption{Our supervised metric Polos computes evaluation scores from multimodal inputs by integrating human feedback within the novel framework $\mathrm{M^2LHF}$. Polos is capable of modeling intricate relationships within the vector space of text-image pairs as well as text-text pairs, thereby effectively evaluating the depicted samples.}
    \label{fig:overview}
    \vspace{-6mm}
\end{figure}

Extensive research on image captioning has yielded a wide array of practical applications in society, spanning from assisting people who are blind to facilitating dialog about images and answering questions derived from visual content (e.g., \cite{blind,open-domain-cqg,capwap}).
To effectively advance the development of image captioning models, it is crucial to establish an automatic evaluation metric that closely aligns with human judgment. 
Previous research has shown that classical automatic evaluation metrics \cite{bleu, meteor, rouge, cider, spice} exhibit weak correlation with human judgments \cite{spice, clipscore}.
This has prompted the introduction of data-driven automatic evaluation metrics \cite{clipscore, bertscore, moverscore,  vilbertscore}.
However, by merely measuring the similarity of embeddings learned from tasks unrelated to image captioning, these metrics potentially misjudge caption quality, which raises concerns about their accuracy in evaluating image captioning models.
Furthermore, some experiments have identified the limitations of these metrics regarding handling \textit{hallucinations} adequately. 

Although recent similarity-based \cite{clipscore, bertscore, moverscore, vilbertscore, mid} and learning-based \cite{umic,pac-s} metrics have demonstrated performance superior to classic metrics, they are still not completely satisfactory. 
For instance, while the correlation between the state-of-the-art (SOTA) metric and human judgments on the Flickr8K dataset \cite{flickr} is only about 0.56, the correlation coefficient among human judgments is approximately 0.73 \cite{spice}.
This discrepancy may emerge because similarity-based metrics merely compute cosine similarity from well-known embeddings (e.g., \cite{clip, bert, xlm}) learned from tasks unrelated to image captioning evaluation, which potentially leads to a misrepresentation of caption quality.
Moreover, many learning-based metrics are specifically for limited settings (e.g., closed-vocabulary settings), and therefore fail to accommodate diverse images and texts. 


In this paper, we propose Polos, a supervised automatic evaluation metric for image captioning models.
Alongside this, we introduce Multimodal Metric Learning from Human Feedback ($\mathrm{M^2LHF}$), a framework used to develop a practical supervised metric for image captioning. 
Fig.\ref{fig:overview} illustrates our proposed metric Polos and the $\mathrm{M^2LHF}$ framework.
As a key contribution, our proposed metric fuses both similarity-based and learning-based approaches.  
Previous similarity-based approaches merely compute scalar similarities using classic methodologies (e.g., cosine similarity and optimal transport), whereas our metric models intricate relationships in the vector space of text-image pairs and text-text pairs. This is achieved through the \textit{parallel feature extraction mechanism} that leverages SimCSE \cite{simcse} and CLIP \cite{clip}, which we provide details for in Section \ref{sec:main-method}.

To train a metric that embodies the aforementioned characteristics through the $\mathrm{M^2LHF}$ approach, we have constructed the Polaris dataset, which contains a diverse range of human judgments.
Compared with existing datasets, Polaris contains a greater diversity of captions and a more extensive range of evaluations. 
Notably, our dataset comprises 131K human judgments, whereas even the largest existing dataset \cite{composite} contains only approximately 12K in total. 
Regarding the total number of evaluators involved, even the CapEval1K dataset \cite{umic}, which has the largest number of evaluators to the best of our knowledge, was assessed by only five evaluators. By contrast, our dataset is distinguished by its assessment by a much higher number of evaluators at 550 evaluators.
Furthermore, our dataset is superior to existing datasets such as \cite{umic, composite, flickr, spice} in terms of the inclusion of diverse captions, which were collected from humans and generated from ten image captioning models, including modern models.

The main contributions of this paper\footnote{Project page: \url{https://yuiga.dev/polos}} are as follows :
\vspace{-2mm}
\begin{itemize}
    \setlength{\parskip}{0.2mm} 
    \setlength{\itemsep}{0.2mm} 
    \item We propose Polos\footnote{\textbf{P}ractical visi\textbf{O}n-and-\textbf{L}anguage evaluati\textbf{O}n metic for image captioning model\textbf{S}}, a supervised automatic evaluation metric for image captioning models.
    \item We introduce $\mathrm{M^2LHF}$, a novel framework used to develop a practical metric for image captioning. 
    \item We introduce a \textit{parallel feature extraction mechanism} that leverages text embeddings \cite{roberta} pretrained with SimCSE and vision-language embeddings \cite{clip}.
    \item We constructed the Polaris dataset, which contains 131,020 human judgments from 550 evaluators.
    \item We achieved SOTA performance on image captioning benchmarks including Composite, Flickr8K-Expert, and Flickr8K-CF, PASCAL-50S, FOIL, and Polaris.
\end{itemize}
\vspace{-5mm}
\section{Related Work}
\vspace{-2mm}
Data-driven metrics can be broadly divided into similarity-based metrics \cite{clipscore, bertscore, vilbertscore, mid} and learning-based metrics \cite{umic,pac-s, comet}.
Similarity-based metrics compute similarities using classic approaches such as cosine similarity and optimal transport in an unsupervised manner, whereas learning-based metrics compute scores in a supervised manner.

\vspace{-3mm}
\paragraph{Standard and similarity-based metrics.}

Standard automatic metrics for evaluating image captioning models include BLEU \cite{bleu}, ROUGE \cite{rouge}, METEOR \cite{meteor}, CIDEr \cite{cider}, and SPICE \cite{spice}, which are primarily based on either $n$-grams or scene graphs. Extensions to these standard metrics, such as CIDEr-R \cite{cider-r} and JaSPICE \cite{jaspice}, have also been proposed. 
Despite their widespread use, in several studies, researchers have highlighted the limitations of these metrics, indicating their suboptimal performance \cite{bleurt, comet, clipscore, vilbertscore, pac-s}. This has led to the emergence of data-driven metrics such as BERTScore \cite{bertscore} and MoverScore \cite{moverscore}. Additionally, there are similar metrics that leverage image features directly, such as \cite{tiger, vilbertscore, clipscore, umic, mid}.

CLIPScore \cite{clipscore} evaluates captions in an unsupervised manner by computing their similarity with embeddings derived from CLIP \cite{clip}. Its distinctive feature is its capacity not only to evaluate based on a reference-with-image manner but also in a reference-free context. Similarly, PAC-S \cite{pac-s} fine-tunes CLIP on \textit{generated} image-text pairs and evaluates captions in the same manner. However, these dependences solely on the computation of cosine similarity between CLIP embeddings could potentially constrain its effectiveness and robustness across diverse scenarios. 

 MID\cite{mid} uses the negative Gaussian cross-mutual information using CLIP features. It was proposed as a bridge between reference-with-image and reference-free metrics and has demonstrated strong performance across multiple image captioning benchmarks. 
However, it is crucial to highlight that MID relies solely on the text embeddings provided by CLIP for processing text data, which could potentially lead to suboptimal evaluation. This concern stems from the fact that CLIP is predominantly trained on a dataset that comprises mainly short noun phrases, which means that it is possibly suboptimal for evaluating the longer sentences typically generated by image captioning models \cite{pac-s}.

\vspace{-3mm}
\paragraph{Learning-based metrics.}

Limited learning-based metrics exist for image captioning models; however multiple learning-based metrics exist in the field of evaluation of text generation \cite{bleurt, ruse, lens, sescore, sescore2, instructscore}, including RUSE \cite{ruse} and COMET \cite{comet}. 
COMET is a metric trained using human judgments that has demonstrated robust performance in evaluating machine translations. COMET comprises both an estimator model that directly predicts human judgments and a ranking model that predicts the quality order of the generated translations.

By contrast, a few learning-based metrics exist that were specifically designed for image captioning \cite{prmcs, umic, qe_ic}. One such metric is UMIC \cite{umic}, which is among the few learning-based metrics tailored for evaluating image captioning models.  
UMIC is a fine-tuned UNITER \cite{uniter} model designed to rank captions against each other using CapEval1K.
However, as we argue later, such \textit{ranking} models have shortcomings when they process multiple potential references, such as varying focal points in captions and subjective variations in expression.
Furthermore, UMIC is only a straightforward fine-tuned UNITER model and it cannot adequately handle diverse images and text, such as those associated with an open-vocabulary setting.

\vspace{-3mm}
\paragraph{Datasets and benchmarks.}

Standard datasets commonly used for the evaluation of image captioning include Flickr8K-Expert, Flickr8K-CF \cite{flickr}, Composite \cite{composite} and PASCAL-50S \cite{cider}. 
The Flickr8K-Expert and Flickr8K-CF datasets comprise a significant amount of human judgments on captions provided by humans.  
However, these datasets do not contain any captions generated by models, which presents an issue from the perspective of the domain gap when using them for training metrics. 
The Composite dataset \cite{composite} encompasses 12K human judgments across images collected from MSCOCO \cite{coco}, Flickr8k \cite{flickr}, and Flickr30k \cite{flickr30k}. Although each image initially contains five references, only one reference was selected for human judgments within the dataset.
The CapEval1k dataset was introduced by \cite{umic} for training automatic evaluation metrics. We note that the CapEval1K dataset has several limitations: it is a closed dataset, uses outdated models such as \cite{att2in, bottom-up, aoanet}, and includes only 1K human judgments. In stark contrast, our Polaris dataset offers significant advantages: it is openly accessible, incorporates inferences from modern models \cite{sat,m2trm,vinvl,grit,blip,git,ofa,blip2}, and includes a substantial 131K human judgments. A meta-analysis of these datasets can be found in Appendix \ref{appendix:dataset-meta}.

Standard datasets for image captioning include MS-COCO, nocaps \cite{nocaps}, Flickr30K, and CC3M \cite{cc3m}. 
The nocaps dataset contains a greater diversity of classes than MS-COCO, which enables a more comprehensive evaluation of image captioning models' ability to generate diverse captions.
Our Polaris dataset is built on inferences from MS-COCO and nocaps to ensure caption diversity.

\section{Methodology}

\begin{figure}[t]
    \centering
    \includegraphics[width=\linewidth]{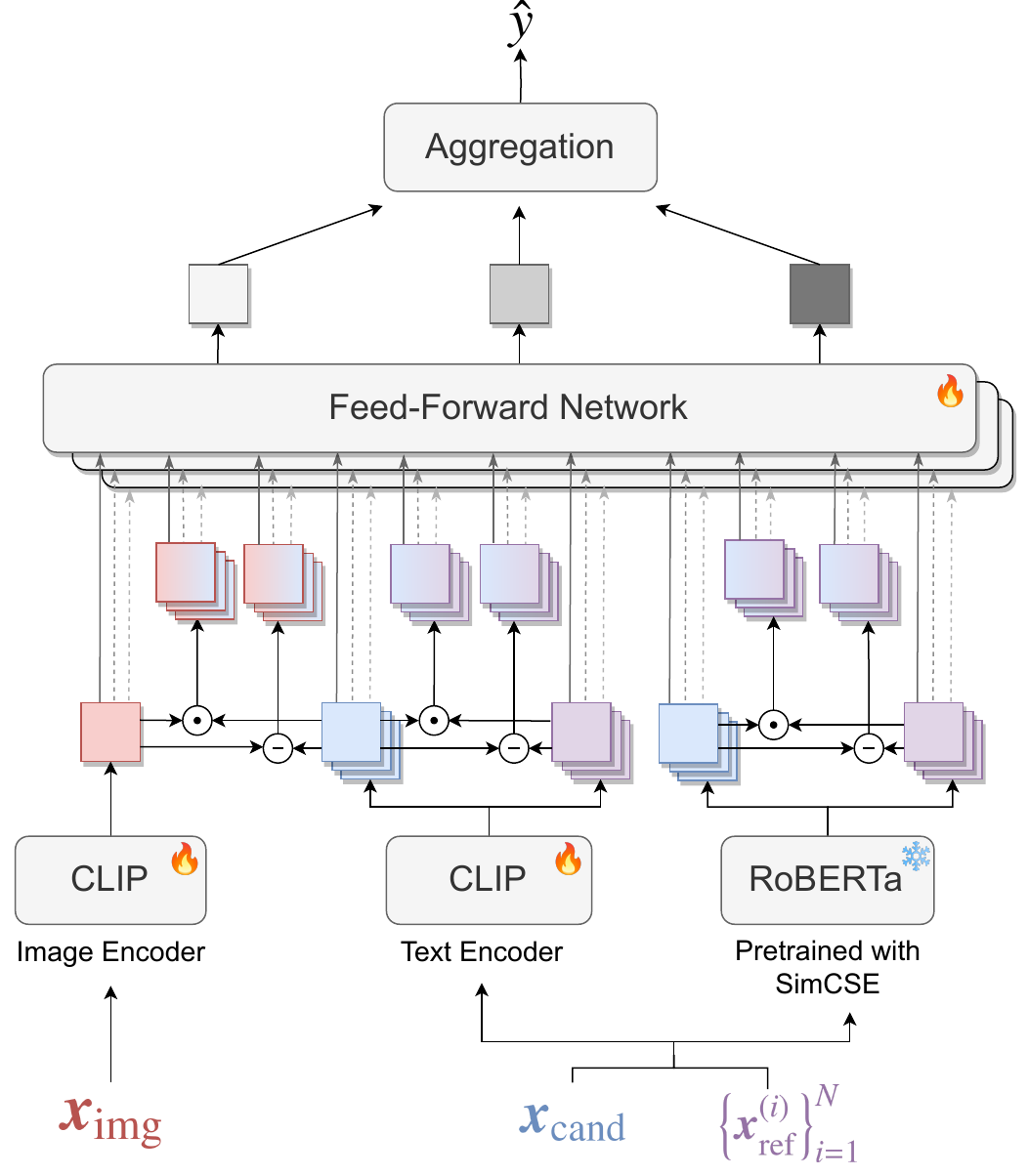}
    \caption{Overview of the proposed metric. In alignment with the principles of $\mathrm{M^2LHF}$, Polos computes the evaluation $\hat{y}$ based on multimodal inputs and regresses the human evaluation. The proposed metric extracts effective features for caption evaluation using the difference and Hadamard product of features derived from both CLIP and RoBERTa.}
    \label{model}
    \vspace{-3mm}
\end{figure}

\subsection{Meta-Analysis}

When designing automatic evaluation metrics, the decision to use an unsupervised or supervised approach is important. 
As detailed in previous research \cite{bartscore}, this binary distinction can be further decomposed into four categories: \textit{regression}, \textit{ranking}, \textit{matching}, and \textit{generation}. We will discuss the above categories in the following subsections.

\vspace{-3mm}
\paragraph{Unsupervised approaches.}
In the domain of unsupervised approaches, metrics such as BERTScore \cite{bertscore}, CLIPScore \cite{clipscore}, and BARTScore \cite{bartscore} are prominent. However, each of these metrics has its limitations for evaluating image captioning models. 
For instance, \textit{matching} models such as BERTScore measure the similarity between tokens, which makes them ill-suited to explicitly handling images. 
Simultaneously, CLIPScore, which computes the similarity between sentence embeddings and image embeddings, has a major drawback. Because CLIP was designed to match an entire image to a text description, it has shortcomings regarding capturing the fine-grained alignment between specific image regions and text spans as indicated in \cite{region-clip, vild}. This limitation suggests that there is potential to improve performance, particularly by transitioning to a supervised approach.
Similarly, unsupervised \textit{generation} models such as BARTScore encounter challenges in the evaluation of image captioning models. This difficulty stems from the limited availability of large-scale, pretrained, and lightweight multimodal encoders.

\vspace{-3mm}
\paragraph{Supervised approaches.}
Based on the above discussion, we believe that supervised metrics have distinct advantages. Both \textit{regression} and \textit{ranking} models are viable options; however, we believe that \textit{regression} models are better suited for evaluating image captioning models that involve multiple references because \textit{ranking} models cannot adequately handle multiple potential references, such as varying areas of interest in captions and inherent subjectivity in expression. 
For instance, the comparison of captions with different focal points lacks substantive meaning and ranking subjective expressions is difficult.
Although COMET \cite{comet} introduces both \textit{ranking} and \textit{regression} models, the task of evaluating machine translation with limited interpretation differs significantly from our task. This distinction is a factor that prevents the straightforward adoption of a \textit{ranking} model.
Given these considerations, in this study, we propose a metric based on \textit{regression}.

\subsection{Metric Design}

\paragraph{Reference-free vs reference-with-image.}
In several studies, researchers have introduced reference-free metrics for image captioning \cite{clipscore, umic,pac-s}. 
However, these metrics have not achieved performance comparable with metrics that consider references. 
Additionally, the performance of reference-free metrics heavily relies on the alignment capabilities between image and language features. 
This is evident from the fact that, on the FOIL dataset, CLIP-S underperforms compared with traditional metrics such as CIDEr \cite{clipscore}. 
Such results suggest potential limitations in handling \textit{hallucinations}, which poses a significant challenge in the field of image captioning. 
Given these considerations, we chose not to adopt the reference-free problem setting.

\vspace{-3mm}
\paragraph{Limitations of CLIP text embeddings.}

As previously highlighted, many data-driven metrics do not have adequate generalization capabilities across diverse image and text types. 
This inadequacy partly stems from the sole use of text embeddings from the CLIP text encoder. CLIP provides versatile textual and visual features, as demonstrated by their performance in various tasks.
However, the CLIP model, pretrained on web-collected image-caption pairs, is likely to be suboptimal for evaluation metrics because these annotations typically lack the richness and descriptiveness necessary for evaluating generated long captions, as indicated in \cite{pac-s}.
Consequently, we posit that sentence embeddings pretrained with supervised SimCSE \cite{simcse} may be a more advantageous approach than CLIP.
This is partially supported by the observation that SimCSE outperformed previous sentence embedding techniques in the semantic text similarity task \cite{agirre-etal-2012-semeval, agirre-etal-2013-sem, agirre-etal-2014-semeval, agirre-etal-2015-semeval, agirre-etal-2016-semeval, cer-etal-2017-semeval, marelli-etal-2014-sick}.

\subsection{Proposed Method: Polos and M\boldmath{$^2$}LHF}
\label{sec:main-method}

We propose Polos, a supervised automatic evaluation metric tailored for image captioning models. Fig.\ref{model} shows the overview of the proposed Polos.
To enhance robustness and practicality, we also present Multimodal Metric Learning from Human Feedback ($\mathrm{M^2LHF}$), a novel framework for developing metrics based on human feedback. 
Within $\mathrm{M^2LHF}$, a metric computes the evaluation $\hat{y}$ based on multimodal input $\bm{x}$ and directly regresses the human evaluation $y$.
Our method is inspired by automatic evaluation metrics for machine translation, such as COMET and BLEURT \cite{bleurt}, which were explicitly designed to predict human judgments. However, our framework differs from previous works \cite{ruse, comet, bleurt} in that it handles both images and text, and learns directly from human judgments based on multimodal inputs.
We consider that our approach, which applies regression using both image and language features, is considered to be broadly applicable to automatic evaluation metrics that have learnable parameters.

We define the input $\bm{x}$ to the model as follows:
\begin{align}
 \bm{x} = \left\{ \bm{x}_\mathrm{cand}, \left\{\bm{x}_\mathrm{ref}^{(i)}\right\}_{i=1}^{N} , \bm{x}_\mathrm{img} \right\}, 
\end{align}
where $\bm{x}_\mathrm{cand} \in \{1, 0\}^{V \times L}$ represents the candidate, $\{\bm{x}_\mathrm{ref}^{(i)}\} \in \{1, 0\}^{N \times V \times L}$ represents the $i$-th reference out of $N$, and $\bm{x}_\mathrm{img} \in \mathbb{R}^{3 \times H \times W}$ represents the image.
Here, $V, L, N, H$ and $W$ denote vocabulary size, maximum token length, number of reference sentences in one sample, and height and width of the image, respectively.

As previously emphasized, selecting an appropriate method for sentence embedding necessitates careful consideration. In this study, we use both the CLIP text encoder and RoBERTa trained with supervised SimCSE to obtain sentence embeddings. 
Initially, using RoBERTa pretrained with supervised SimCSE, we extract sentence embedding $\bm{c}_\mathrm{rb} \in \mathbb{R}^{L \times d_{R}}$ and $\{\bm{r}_\mathrm{rb}^{(i)}\}_{i=1}^N \in \mathbb{R}^{N \times L \times d_{R}}$ from $\bm{x}_\mathrm{cand}$ and $\{\bm{x}_\mathrm{ref}^{(i)}\}_{i=1}^N$, respectively. Note that $d_{R}$ represents the output dimension of RoBERTa and the sentence embeddings are derived from the \verb|[CLS]| token of inputs into RoBERTa.
Following this, using the CLIP text encoder, we derive language feature $\bm{c}_\mathrm{clip} \in \mathbb{R}^{d_\mathrm{CLIP}}$ and $\bm{r}_\mathrm{clip}^{(i)} \in \mathbb{R}^{d_\mathrm{CLIP}}$ from $\bm{x}_\mathrm{cand}$ and $\bm{x}_\mathrm{ref}^{(i)}$, respectively, where $d_\mathrm{CLIP}$ denotes the output dimension of the CLIP encoders.
Additionally, utilizing the pretrained CLIP image encoder (ViT-B/16), we obtain image features $\bm{v} \in \mathbb{R}^{d_\mathrm{CLIP}}$ from $\bm{x}_\mathrm{img}$.


\vspace{-3mm}
\subsubsection{Parallel Feature Extraction Mechanism}
In alignment with the principles of $\mathrm{M^2LHF}$, we also propose a \textit{parallel feature extraction mechanism}. This mechanism serves as a multimodal extension of the RUSE method\cite{ruse, comet}, employing both difference and Hadamard products. 
It extracts effective features for caption evaluation by utilizing the difference and Hadamard product of features derived from both CLIP and RoBERTa. 
Given that CLIP is designed to minimize the cosine similarity between corresponding language and image features, the Hadamard product applied to CLIP features is considered to be effective.
Additionally, the difference and Hadamard product operations generate vectors that encapsulate similarity because each element of the vector can be amplified or attenuated in relation to the others.

Initially, given the following inputs: 
\begin{align}
\left\{\bm{c}_\mathrm{clip}, \: \bm{r}_\mathrm{clip}^{(i)}, \: \bm{c}_\mathrm{rb}, \: \bm{r}_\mathrm{rb}^{(i)}, \: \bm{v} \right\},
\end{align}
the proposed framework first computes $\bm{h}_\mathrm{inter}^{(i)}$ as:
\begin{align}
\bm{h}_\mathrm{inter}^{(i)} = [F(\bm{c}_\mathrm{clip}, \bm{r}_\mathrm{clip}^{(i)}); \: F(\bm{c}_\mathrm{clip}, \bm{v}); \: F(\bm{c}_\mathrm{rb}, \bm{r}_\mathrm{rb}^{(i)})].
\end{align}
Here, $F$ is the following function:
\begin{align}
\label{eq:F}
F(\bm{c},\bm{r}) = \left[\bm{c}; \: \bm{r}; \: |\bm{c} - \bm{r}|; \: \bm{c} \odot \bm{r}\right],
\end{align}
where $\odot$ denotes the Hadamard product.

Subsequently, we apply the MLP to $\bm{h}_\mathrm{inter}^{(i)}$ to compute $\bm{h}^{(i)}$, which effectively captures the similarity among the multimodal features for the $i$-th reference.
Note that we chose MLP because pilot experiments demonstrated its superior performance compared with Transformer \cite{transformer}.

Finally, we compute the evaluation score $\hat{y}$ as:
\begin{align}
\hat{y} = \underset{i}{\mathrm{Aggregate}}(\sigma(\mathrm{MLP}(\bm{h}^{(i)}))),
\end{align}
where $\sigma$ denotes the sigmoid function, which scales the output to the range $[0, 1]$.
In this context, $\mathrm{Aggregate}$ denotes an aggregation function. This function can encompass various operations, such as calculating the maximum or average value. Notably, in our experimental setup, we opted for the max function.

For the loss function, we adopted the mean squared error, which is a standard choice in regression problems because of its effectiveness in quantifying the variance between predicted and human judgments. Our implementation details can be found in Appendix \ref{appendix:impl}.
\section{Experimental Evaluation}

\subsection{Setups}

\paragraph{Polaris dataset.}
In this study, we introduce the Polaris dataset, which consists of image-caption pairs and human judgments on the appropriateness of the captions.
Training supervised models to predict human judgments benefits significantly from a large-scale corpus that contains diverse captions.
However, to the best of our knowledge, there are few open datasets with diverse captions.
Therefore, we constructed the Polaris dataset, which contains a total of 131,020 human judgments collected from 550 evaluators.

In comparison with existing datasets, which comprise the inference results of image captioning models and are suitable for metric training, Polaris is distinguished by its exceptional diversity of captions and broader spectrum of evaluations. Our dataset, which comprises over 131K human judgments, significantly surpasses the largest dataset suitable for metric training, which contains only around 12K judgments \cite{composite}.
Regarding the diversity of evaluations, the Polaris dataset had an average of eight evaluators per caption, which provides a more comprehensive judgment than the Flickr8K dataset, which had an average of only three evaluators per caption. 
 Even the CapEval1K dataset, one of the largest in terms of evaluator participation, only involved five evaluators, which underscores the extensive nature of Polaris with its inclusion of 550 evaluators.
Moreover, Polaris differentiates itself from existing datasets such as \cite{umic, composite, flickr, spice} by being openly accessible, constructed with modern models, and incorporating a wide range of diverse captions.

The Polaris dataset comprises captions generated by the ten standard models: $\mathrm{SAT}$ \cite{sat}, $\mathrm{\mathcal{M}^2}$-$\mathrm{Transformer}$ \cite{m2trm}, $\mathrm{VinVL}$ \cite{vinvl}, $\mathrm{GRIT}$ \cite{grit}, $\mathrm{BLIP_\mathrm{base}}$, $\mathrm{BLIP_\mathrm{large}}$ \cite{blip}, $\mathrm{GIT}$ \cite{git}, $\mathrm{OFA}$ \cite{ofa}, $\mathrm{BLIP}$-$\mathrm{2_\mathrm{flan}}$, and $\mathrm{BLIP}$-$\mathrm{2_\mathrm{opt}}$ \cite{blip2}. Within this set, $\mathrm{BLIP_\mathrm{base}}$ and $\mathrm{BLIP_\mathrm{large}}$ refer to versions of BLIP that employ ViT-B and ViT-L as their image encoders. Similarly, $\mathrm{BLIP}$-$\mathrm{2_\mathrm{flan}}$ and $\mathrm{BLIP}$-$\mathrm{2_\mathrm{opt}}$ denote versions of BLIP-2 that adopt Flan-T5 \cite{flant5} and OPT \cite{opt} as their Large Language Models (LLMs), respectively.
We selected these models because they are standard image captioning models. 
Additionally, we also chose older models to ensure diversity in the quality of their output sentences. 
We included inference results for each model, as performed on the MS-COCO \cite{coco} and nocaps \cite{nocaps} datasets in the Polaris dataset. We selected MS-COCO because it is the standard dataset for image captioning, whereas we chose nocaps because of its greater diversity of classes compared with MS-COCO.

For a given image, human evaluators assessed the appropriateness of its caption using a five-point scale, taking into account factors such as fluency, relevance, and descriptiveness. We used a crowdsourcing service to collect these evaluations.
In the Polaris dataset, we transformed the human judgments, which were rated on a five-point scale, to values in the range $[0,1]$ using min-max normalization.
To eliminate unreliable data, we excluded data from evaluators who exhibited suspicious behavior, such as extremely short response times or consistently providing identical values. The statistical information and details of the Polaris dataset can be found in Appendix \ref{appendix:stat}.

\vspace{-3mm}
\paragraph{Baseline metrics.}

We adopted BLEU \cite{bleu}, ROUGE \cite{rouge}, METEOR \cite{meteor}, CIDEr \cite{cider} and SPICE \cite{spice} because they are standard metrics for image captioning tasks. Additionally, we included MoverScore \cite{moverscore}, BERTScore \cite{bertscore}, BARTScore \cite{bartscore}, ViLBERTScore \cite{vilbertscore}, TIGEr \cite{tiger}, LEIC \cite{leic}, CLIPScore \cite{clipscore}, MID \cite{mid}, UMIC \cite{umic} and PAC-S \cite{pac-s} as baseline metrics because they are representative metrics for image captioning.

\vspace{-3mm}
\paragraph{Benchmarks.}
To assess the practicality of a supervised metric, it is essential to evaluate the metric using both in-domain and out-of-domain datasets.
Particularly in the context of supervised automatic evaluation, cases exist in which supervised metrics seemingly outperform unsupervised metrics on test sets (in-domain).
However, as we demonstrate in the following section, this does not inherently imply better performance on out-of-domain data. 
Given that supervised metrics are frequently applied to out-of-domain data, those that lack robustness can be impractical.
Therefore, evaluating zero-shot performance in supervised models is paramount. In this study, in addition to the Polaris dataset, we used Composite, Flickr8K, PASCAL-50S and FOIL to evaluate zero-shot performance.

\subsection{Correlation with Human Judgments}

\subsubsection{Caption-level Likert judgments}
\label{sec:caption-level-exp}

\begin{table}[t]
    \centering
    \normalsize
    \setlength{\tabcolsep}{5pt}
    \scalebox{0.9}{
    \begin{tabular}{
    >{\raggedright\arraybackslash}p{29mm}
    >{\centering\arraybackslash}p{12mm}
    >{\centering\arraybackslash}p{12mm}
    >{\centering\arraybackslash}p{12mm}
    >{\centering\arraybackslash}p{10mm}
    }
    \toprule
    {} & {\small Composite} & {\small \begin{tabular}{c}Flickr8K\\(Expert)\end{tabular}} & {\small \begin{tabular}{c}Flickr8K\\(CF)\end{tabular}} & {\small Polaris} \\  \hline
\rowcolor{TitleColor} \multicolumn{2}{l}{Classic metrics} & {} & {} & {} \\ 
{BLEU \cite{bleu}} & {30.6} & {30.8} & {16.4} & {46.3} \\ 
{ROUGE \cite{rouge}} & {32.4} & {32.3} & {19.9} & {46.3} \\
{CIDEr \cite{cider}} & {37.7} & {43.9} & {24.6} & {52.1} \\
{METEOR \cite{meteor}} & {38.9} & {41.8} & {22.2} & {51.2} \\
{SPICE \cite{spice}} & {40.3} & {44.9} & {24.4} & {51.0} \\
{SPARCS \cite{smurf}} & {43.1} & {48.1} & {10.4} & {43.3} \\ \hline
\rowcolor{TitleColor} \multicolumn{2}{l}{Similarity-based metrics} & {} & {} & {} \\ 
{MoverScore \cite{moverscore}} & {30.1} & {46.7} & {22.8} & {46.4} \\
{BERTScore \cite{bertscore}} & {30.1} & {46.7} & {22.8} & {51.6} \\
{BARTScore \cite{bartscore}} & {43.5} & {37.8} & {24.3} & {47.3} \\
{LEIC \cite{leic}} & {--} & {--} & {29.5} & {--} \\
{TIGEr \cite{tiger}} & {45.4} & {--} & {--} & {--} \\
{ViLBERTScore \cite{vilbertscore}} & {52.4} & {50.1} & {--} & {--} \\
{CLIP-S \cite{clipscore}} & {53.8} & {51.2} & {34.4} & {52.3} \\
{RefCLIP-S \cite{clipscore}} & {55.4} & {53.0} & {36.4} & {52.3} \\
{MID \cite{mid}} & {55.7} & {54.9} & {37.3} & {51.3} \\ \hline
\rowcolor{TitleColor} \multicolumn{2}{l}{Learning-based metrics} & {} & {} & {} \\
{PAC-S \cite{pac-s}} & {55.7} & {54.3} & {36.0} & {52.5} \\  
{UMIC \cite{umic}} & {56.1} & {46.8} & {30.1} & {49.8} \\
{RefPAC-S \cite{pac-s}} & \underline{{57.3}} & \underline{{55.9}} & \underline{{37.6}} & \underline{{56.0}} \\  \hline
\rowcolor{LightCyan}
{} & \textbf{57.6} & \textbf{56.4} & \textbf{37.8} & \textbf{57.8} \\ 
\rowcolor{LightCyan}
\multirow{-2}{*}{\textbf{Polos (Ours)}} & (\textcolor{blue}{+0.3}) & (\textcolor{blue}{+0.5}) & (\textcolor{blue}{+0.2}) & (\textcolor{blue}{+1.8}) \\ 
\bottomrule
    \end{tabular}
    }
    \caption{Correlation coefficients between various metrics and human judgments. The symbol `--' indicates non-executable code or unavailable data. \textbf{Bold} font indicates the highest recorded value and \underline{{underlining}} indicates the second-highest value.}
    \vspace{-5mm}
    \label{table:subset-res}
\end{table}

\begin{figure*}[t]
    \centering
    \includegraphics[width=\linewidth]{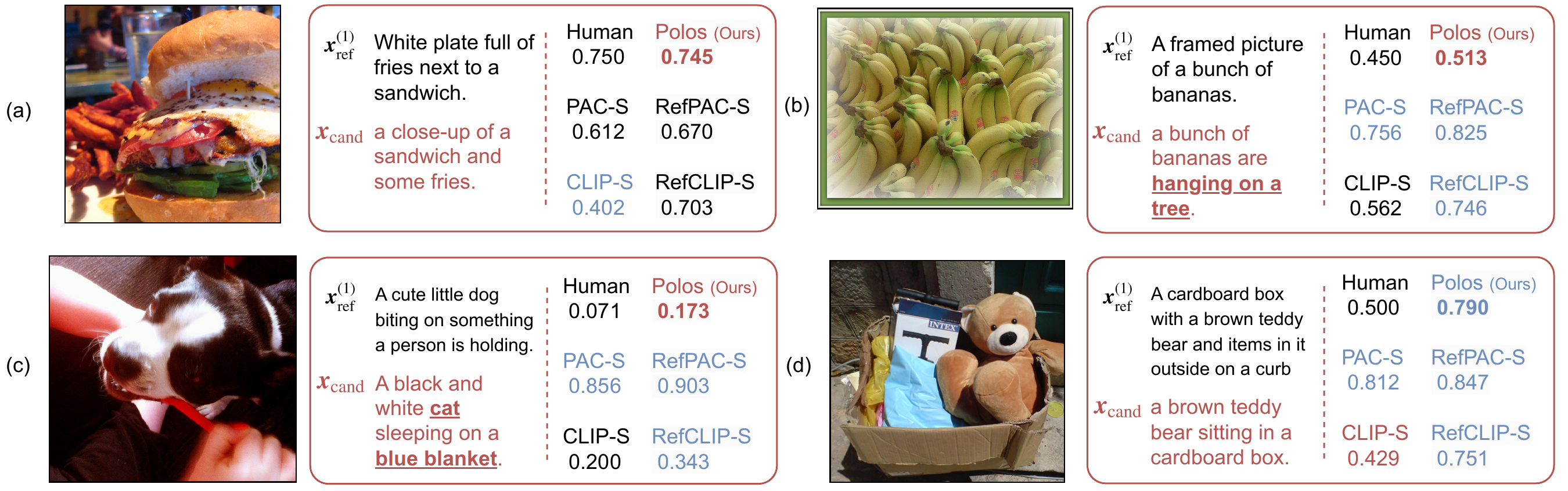}
    \caption{Examples of successful and failed cases from the Polaris dataset. Values in \textbf{\textcolor{myblue}{blue}} indicate critical errors, and values in \textbf{\textcolor{myred}{red}} represent those closest to the human judgments. \underline{Underlined} words indicate significant inaccuracies in the candidate captions. These results demonstrate that our proposed metric effectively handled multimodal inputs and yielded evaluation scores that aligned closely with human judgment. Note that $\bm{x}_\mathrm{ref}^{(1)}$ and $\bm{x}_\mathrm{cand}$ denote one of the reference captions and the candidate caption, respectively.}
    \label{fig:qualitive}
    \vspace{-5mm}
\end{figure*}

We conducted comparative experiments with baselines across Composite, Flickr8K-Expert, Flickr8K-CF, and the Polaris dataset.
The results are presented in Table \ref{table:subset-res}, which details the comparison of Kendall’s $\tau$ with the baselines across the aforementioned datasets. Following the methodology of previous research \cite{clipscore, mid, spice}, we used $\tau_b$ (Kendall-B) for the Flickr8K-CF dataset and $\tau_c$ (Kendall-C) for all other datasets.
Notably, our proposed metric achieved SOTA results with scores of 57.6, 56.4, 37.8, and 57.8 for Composite, Flickr8K-Expert, Flickr8K-CF and the Polaris dataset, respectively.
Specifically, our metric outperformed RefPAC-S by margins of 0.3, 0.5, 0.2, and 1.8 points on Composite, Flickr8K-Expert, Flickr8K-CF, and the Polaris dataset, respectively. This indicates that our supervised metric is superior to other supervised metrics such as RefPAC-S.
Moreover, we achieved better performance than reference-free metrics such as PAC-S and intermediary metrics (metrics positioned between reference-with-image and reference-free metrics) such as MID. This highlights the critical role of incorporating references.

Fig.\ref{fig:qualitive} shows various examples of the proposed metric for the Polaris dataset (further results are provided in Appendix \ref{appendix:qualitive}).
First, Fig.\ref{fig:qualitive} (a) shows the sample that human evaluators deemed high quality. In this sample, $\bm{x}_\mathrm{cand}$ captured the image content suitably, which resulted in a high human judgment of 0.750. CLIP-S assessed it at 0.402, whereas our proposed method aligned more closely with the human evaluation with a score of 0.745.
Fig.\ref{fig:qualitive} (b) illustrates samples rated as mediocre by human evaluators because of partial accuracy. For instance, the veracity of the phrase ``hanging on a tree'' could not be confirmed by the image content, which led to a moderate human score of 0.450. RefPAC-S and RefCLIP-S overestimated the caption with scores of 0.825 and 0.746, respectively, whereas our proposed metric provided a more judicious score of 0.513, thereby reflecting its effectiveness in recognizing ``normal'' quality captions.
Fig.\ref{fig:qualitive} (c) shows examples labeled as poor by human evaluators. A discrepancy, such as misidentifying the animal in the reference, resulted in a low human score of 0.071. Again, RefPAC-S and PAC-S assessments overestimated the caption, with scores at 0.903 and 0.856, respectively, whereas our method assigned a more realistic score of 0.173, effectively capturing the \textit{hallucination} error.
These results show that our proposed metric appropriately handled multimodalities and produced evaluation scores close to human judgments.

Fig.\ref{fig:qualitive} (d) illustrates a sample for which the proposed metric did not perform as expected. 
In Fig.\ref{fig:qualitive} (d), $\bm{x}_\mathrm{ref}^{(1)}$ and $\bm{x}_\mathrm{ref}^{(2)}$ were described as ``A cardboard box with a brown teddy bear and items in it outside on a curb'' and ``a box full of stuffed animals and other children's items,'' respectively.
Conversely, $\bm{x}_\mathrm{cand}$ was described as ``a brown teddy bear sitting in a cardboard box.''
For this sample, the average of human judgments was $\bm{x}_\mathrm{cand}$ as 0.50.
We believe that this was because the box contained various items in addition to the teddy bear.
By contrast, our proposed metric evaluated this sample with a score of 0.790, which indicates a notable disparity from human judgment.
Similarly, RefPAC-S and RefCLIP-S output scores of 0.847 and 0.751, respectively, which also demonstrates a discrepancy between them and human judgment. 
Given the failure of our metric and the CLIPScore family, the primary cause of these failures is likely to be an overemphasis on objects that are prominently visible, which results in overlooking the broader context of the image. This is likely to be caused by CLIP's shortcomings in capturing the fine-grained alignment between specific image regions and text, as indicated in \cite{region-clip, vild}.

\vspace{-3mm}
\subsubsection{Pairwise ranking on Pascal-50S}

\begin{table}[t]
    \centering
    \normalsize
    \scalebox{0.9}{
    \begin{tabular}{lccccc}
    \toprule
{} & {HC} & {HI} & {HM} & {MM} & {Mean} \\ \hline
\rowcolor{TitleColor} \multicolumn{2}{l}{Classic metrics} & {} & {} & {} & {} \\ 
{BLEU \cite{bleu}} & {60.4} & {90.6} & {84.9} & {54.7} & {72.7} \\
{METEOR \cite{meteor}} & {63.8} & {97.7} & {93.7} & {65.4} & {80.2} \\
{ROUGE \cite{rouge}} & {63.7} & {95.3} & {92.3} & {61.2} & {78.1} \\
{SPICE \cite{spice}} & {63.6} & {96.3} & {86.7} & {68.3} & {78.7} \\
{CIDEr \cite{cider}} & {65.1} & {98.1} & {90.5} & {64.8} & {79.6} \\ \hline
\rowcolor{TitleColor} \multicolumn{2}{l}{Similarity-based metrics} & {} & {} & {} & {} \\ 
{ViLBERTScore \cite{vilbertscore}} & {49.9} & {99.6} & {93.1} & {75.8} & {79.6} \\
{BERTScore \cite{bertscore}} & {65.4} & {98.1} & {96.4} & {60.3} & {80.1} \\
{MoverScore \cite{moverscore}} & {65.1} & {97.1} & {93.2} & {65.6} & {80.3} \\
{TIGEr \cite{tiger}} & {56.0} & \textbf{99.8} & {92.8} & {74.2} & {80.7} \\
{CLIP-S \cite{clipscore}} & {56.5} & {99.3} & {96.4} & {70.4} & {80.7} \\
{RefCLIP-S \cite{clipscore}} & {64.5} & {99.6} & {95.4} & {72.8} & {83.1} \\
{MID \cite{mid}} & {67.0} & \underline{{99.7}} & \underline{{97.4}} & \underline{{76.8}} & \underline{{85.2}} \\ \hline
\rowcolor{TitleColor} \multicolumn{2}{l}{Learning-based metrics} & {} & {} & {} & {} \\ 
{PAC-S \cite{pac-s}} & {60.6} & {99.3} & {96.9} & {72.9} & {82.4} \\ 
{RefPAC-S \cite{pac-s}} & \underline{{67.7}} & {99.6} & {96.0} & {75.6} & {84.7} \\ 
{UMIC \cite{umic}} & {66.1} & \textbf{99.8} & \textbf{98.1} & {76.2} & {85.1} \\ \hline
\rowcolor{LightCyan}
{} & \textbf{70.0} & {99.6} & \underline{{97.4}} & \textbf{79.0} & \textbf{86.5} \\ 
\rowcolor{LightCyan}
\multirow{-2}{*}{\textbf{Polos (Ours)}} & (\textcolor{blue}{+3.0}) & {} & {} & (\textcolor{blue}{+1.2}) & (\textcolor{blue}{+1.3}) \\ \bottomrule
    \end{tabular}
    }
    \caption{Pascal50-S accuracy results (five references). }
    \vspace{-5mm}
    \label{table:pascal}
\end{table}

Pascal-50S \cite{cider} introduced an alternative evaluation framework for accuracy, which comprised 4K pairwise preference judgments between two captions. These judgments encompass four distinct scenarios: pairs of HC (human correct) captions, HI pairs (both human-written, with one incorrect), HM pairs (one from a human and the other generated by a machine), and MM pairs (both generated by machines). The caption that received the majority vote was deemed the preferred choice, with ties resolved randomly. In our study, following the procedure in \cite{clipscore}, we calculated the average over five evaluations randomly selected from 48 candidates. 
Table \ref{table:pascal} shows the accuracy results for PASCAL-50S. In our experiments, we achieved SOTA results with accuracies of 70.0\%, 79.0\%, and 86.5\% for HC, MM, and Mean, respectively.
The results in Tables \ref{table:subset-res} and \ref{table:pascal} demonstrate that our proposed metric outperformed the baselines in terms of zero-shot performance.
These results provide strong evidence that our metric is both robust and practical, thereby serving as a potent automatic evaluation metric for image captioning.

\vspace{-3mm}
\subsubsection{Sensitivity to hallucination}
\begingroup
\setlength{\columnsep}{5pt}
\captionsetup[table]{skip=3pt}
\begin{wraptable}{r}{4.5cm}
    \vspace{-3mm}
    \centering
    \normalsize
    \scalebox{0.86}{
    \begin{tabular}{lcc}
    \toprule
{} & {1-ref} & {4-ref} \\ \hline
{BLEU \cite{bleu}} & {66.5} & {82.6} \\
{ROUGE \cite{rouge}} & {71.7} & {79.3} \\
{METEOR \cite{meteor}} & {78.8} & {82.6} \\
{CIDEr \cite{cider}} & {82.5} & {90.6} \\
{SPICE \cite{spice}} & {75.5} & {86.1} \\ \hline
{BARTScore \cite{bartscore}} & {85.3} & {91.1} \\
{MoverScore \cite{moverscore}} & {88.4} & {88.4} \\
{BERTScore \cite{bertscore}} & {88.6} & {92.1} \\ \hline
{CLIP-S \cite{clipscore}} & {87.2} & {87.2} \\
{MID \cite{mid}} & {90.5} & {90.5} \\ \hline
{PAC-S \cite{pac-s}} & {89.9} & {89.9} \\
{RefCLIP-S \cite{clipscore}} & {91.0} & {92.6} \\
{RefPAC-S \cite{pac-s}} & \textbf{93.7} & \underline{{94.9}} \\ \hline
\rowcolor{LightCyan}
{\textbf{Polos (Ours)}} & \underline{{93.3}} & \textbf{95.4} \\
\bottomrule
    \end{tabular}
    \vspace{-5mm}
    }
    \caption{FOIL hallucination pairwise detection accuracy results.}
    \label{table:foil}
\end{wraptable}
Previous studies \cite{clipscore, mid} measured how evaluation metrics handle \textit{hallucinations} in captions using the FOIL (Find One mismatch between Image and Language caption) dataset \cite{foil}. 
Following the procedure used in \cite{clipscore}, we evaluated 32K test images with either one or four references. Subsequently, we computed the accuracy of various metrics to evaluate their effectiveness in consistently awarding higher scores to the true candidate compared with the FOIL dataset.

Table \ref{table:foil} presents the accuracy results for the FOIL dataset. 
Our proposed method outperformed previous metrics, achieving SOTA results in the 4-ref setting.
Specifically, it achieved an accuracy of 93.3\% in the 1-ref setting and 95.4\% in the 4-ref setting.
As previously mentioned, CLIP-S lags behind the traditional metric CIDEr, whereas our method outperformed it in both the 1-ref and 4-ref settings. 
Moreover, compared with reference-with-image metric RefPAC-S as well as the intermediary metric MID, our approach led by 0.6 and 4.9 points in the 4-ref setting.

\endgroup

\vspace{-1mm}
\subsection{Ablation Study}
\vspace{-1mm}

We conducted three ablation studies to demonstrate the effectiveness of our proposed method.
Table \ref{table:ablation} presents the results of the ablation studies.

\vspace{-3mm}
\paragraph{Parallel feature extraction ablation.}
We investigated the performance of our \textit{parallel feature extraction mechanism} by excluding the Hadamard product and difference, specifically by modifying $F(\mathbf{c}, \mathbf{r})$ (eq.\ref{eq:F}) to the function $F'(\mathbf{c}, \mathbf{r}) = [\mathbf{c}; \: \mathbf{r}]$. 
The correlation coefficients between Metric (i) and human judgments were lower by 18.3, 15.4, and 6.4 points compared to Metric (vi) for the Composite, Flickr8K, and Polaris, respectively. These indicate that our \textit{parallel feature extraction mechanism} provided superior performance.

\vspace{-3mm}
\paragraph{$\bf{M^2LHF}\:$ablation.}
We investigated the performance of $\mathrm{M^2LHF}$ by modifying our model to make predictions based solely on text, excluding any of the following: $\bm{x}_\mathrm{img}$, $(\bm{c}_\mathrm{clip},  \{\bm{r}_\mathrm{clip}^{(i)}\}_{i=1}^N)$, or $(\bm{c}_\mathrm{rb},  \{\bm{r}_\mathrm{rb}^{(i)}\}_{i=1}^N)$.
Initially, by omitting $\bm{x}_\mathrm{img}$, we assessed the significance of the image feature. Compared to Metric (vi), the correlation coefficients between Metric (ii) and human judgments were lower by 0.4, 1.4, and 0.7 points on Composite, Flickr8K, and Polaris respectively. These results suggest that the image feature played a pivotal role in enhancing the performance of our proposed metric.
Subsequently, we evaluated the contribution of each module by excluding either CLIP or RoBERTa. 
Relative to Metric (vi), the correlation coefficients between Metric (iii) and human judgments decreased by 2.6, 3.2, and 2.4 points. Similarly, the exclusion of RoBERTa led to a decrease in performance.
Although we observed that RoBERTa pretrained by SimCSE was found to enhance performance, these results highlighted CLIP as the most influential feature extractor.
Overall, these results validated the efficacy of the image feature and each module, underscoring that the introduction of $\mathrm{M^2LHF}$ contributed to the performance improvement of the proposed metric.

\vspace{-3mm}
\paragraph{Aggregation mechanism ablation.}
We investigated the impact on performance by setting the $\mathrm{Aggregate}$ function to either $\mathrm{Max}$ or $\mathrm{Mean}$.
The correlation coefficients between Metric (v) and human judgments were lower by 2.5, 1.0, and 5.7 points compared with Metric (vi) for the Composite, Flickr8K, and the Polaris dataset, respectively. These results indicate that using the $\mathrm{Max}$ function as the $\mathrm{Aggregate}$ function provided superior performance.


\begin{table}[t]
    \centering
    \normalsize
    \setlength{\tabcolsep}{2pt}
    \scalebox{0.8}{
    \begin{tabular}{ccccccccc}
    \toprule
{\small Metric} & {\small \textit{P}} & {\small $\bm{x}_\mathrm{img}$} & {\small CLIP} & {\small RoBERTa} & {\small $\mathrm{Aggregate}$} & {\small Composite} & {\small Flickr8K} & {\small Polaris}  \\ \hline
{(i)} & {} & {$\checkmark$} & {$\checkmark$} & {$\checkmark$} & $\mathrm{Max}$ & {39.3} & {41.0} & {51.4} \\
{(ii)} & {$\checkmark$} & {} & {$\checkmark$} & {$\checkmark$} & $\mathrm{Max}$ & {56.8} & {55.5} & {57.1} \\
{(iii)} & {$\checkmark$} & {} & {} & {$\checkmark$} & $\mathrm{Max}$ & {55.0} & {53.2} & {55.4} \\
{(iv)} & {$\checkmark$} & {$\checkmark$} & {$\checkmark$} & {} & $\mathrm{Max}$ & {56.0} & {55.0} & {57.4} \\
{(v)} & {$\checkmark$} & {$\checkmark$} & {$\checkmark$} & {$\checkmark$} & $\mathrm{Mean}$ & {55.1} & {55.4} & {52.1} \\ 
\rowcolor{LightCyan}
{(vi)} & {$\checkmark$} & {$\checkmark$} & {$\checkmark$} & {$\checkmark$} & $\mathrm{Max}$ & \textbf{57.6} & \textbf{56.4} & \textbf{57.8} \\ \bottomrule
    \end{tabular}
    }
    \caption{Ablation study results: performance of $\mathrm{M^2LHF}$ and the \textit{parallel feature extraction} and comparison of the effects of using either the $\mathrm{Max}$ or $\mathrm{Mean}$ within the $\mathrm{Aggregate}$ function. `\textit{P}' indicates the use of the \textit{parallel feature extraction mechanism}}
    \label{table:ablation}
    \vspace{-5mm}
\end{table}

\vspace{-2mm}
\subsection{Discussion and Limitations}
\label{sec:discussions}

Although our metric generated compelling results, our approach has some limitations.
As shown in Fig.\ref{fig:qualitive}, our metric demonstrated a tendency to overestimate captions that lacked intricate details. This phenomenon is likely to be attributed to its excessive focus on prominently visible objects; hence it consequently overlooked the broader context within the image.
As discussed in Section \ref{sec:caption-level-exp}, this limitation may be attributed to CLIP's inherent limitations in capturing the fine-grained alignment between specific image regions and their corresponding textual descriptions.
Despite this, we firmly believe that this study represents a significant stride toward the development of a more practical metric for image captioning models.
In future work, we plan to extend our metric by enhancing the fine-grained alignment, drawing inspiration from methods such as RegionCLIP \cite{region-clip}.


\vspace{-2mm}
\section{Conclusion}
\vspace{-2mm}

In this paper, we introduced Polos, an automatic evaluation metric for image captioning. 
The contribution of this paper is fourfold: i) the introduction of $\mathrm{M^2LHF}$, a novel framework used to develop a practical metric for image captioning; ii) the introduction of a \textit{parallel feature extraction mechanism} that leverages CLIP and RoBERTa pretrained with SimCSE; iii) the construction of the Polaris dataset, containing a total of 131,020 human judgments collected from 550 evaluators; and iv) achieving SOTA performance on image captioning benchmarks including Composite, Flickr8K-Expert, and Flickr8K-CF, PASCAL-50S, FOIL, and the Polaris dataset.


\vspace{-2mm}
\section*{Acknowledgments}
This work was supported by a grant from Apple, Inc. Any views, opinions, findings, and conclusions or recommendations expressed in this material are those of the authors and should not be interpreted as reflecting the views, policies or position, either expressed or implied, of Apple, Inc. 
This work was also partially supported by JSPS KAKENHI Grant Number 23H03478, JST CREST, and NEDO.

{\small
\bibliographystyle{ieee_fullname}
\bibliography{reference}

\begin{thebibliography}{10}\itemsep=-1pt

\bibitem{composite}
Somak Aditya, Yezhou Yang, Chitta Baral, Cornelia Fermuller, et~al.
\newblock {From Images to Sentences through Scene Description Graphs using Commonsense Reasoning and Knowledge}.
\newblock {\em arXiv preprint arXiv:1511.03292}, 2015.

\bibitem{agirre-etal-2014-semeval}
Eneko Agirre, Carmen Banea, Claire Cardie, Daniel Cer, and Mona Diab.
\newblock {SemEval-2014 Task 10: Multilingual Semantic Textual Similarity}.
\newblock In {\em {S}em{E}val}, pages 81--91, 2014.

\bibitem{agirre-etal-2015-semeval}
Eneko Agirre, Carmen Banea, et~al.
\newblock {SemEval-2015 Task 2: Semantic Textual Similarity, English, Spanish and Pilot on Interpretability}.
\newblock In {\em {S}em{E}val}, pages 252--263, 2015.

\bibitem{agirre-etal-2016-semeval}
Eneko Agirre, Carmen Banea, et~al.
\newblock {SemEval-2016 Task 1: Semantic Textual Similarity, Monolingual and Cross-Lingual Evaluation}.
\newblock In {\em {S}em{E}val}, pages 497--511, 2016.

\bibitem{agirre-etal-2012-semeval}
Eneko Agirre, Daniel Cer, Mona Diab, and Aitor Agirre.
\newblock {SemEval-2012 Task 6: A Pilot on Semantic Textual Similarity}.
\newblock In {\em {S}em{E}val}, pages 385--393, 2012.

\bibitem{agirre-etal-2013-sem}
Eneko Agirre, Daniel Cer, Mona Diab, Aitor Agirre, and Weiwei Guo.
\newblock {SEM 2013 Shared Task: Semantic Textual Similarity}.
\newblock In {\em SEM}, pages 32--43, 2013.

\bibitem{nocaps}
Harsh Agrawal, Karan Desai, et~al.
\newblock {nocaps: Novel Object Captioning at Scale}.
\newblock In {\em ICCV}, pages 8948--8957, 2019.

\bibitem{blind2}
Hiba Ahsan, Daivat Bhatt, Kaivan Shah, and Nikita Bhalla.
\newblock {Multi-Modal Image Captioning for the Visually Impaired}.
\newblock In {\em NAACL-HLT}, pages 53--60, 2021.

\bibitem{flamingo}
Jean Alayrac, Jeff Donahue, Pauline Luc, et~al.
\newblock {Flamingo: A Visual Language Model for Few-shot Learning}.
\newblock In {\em NeurIPS}, volume~35, pages 23716--23736, 2022.

\bibitem{spice}
Peter Anderson, Basura Fernando, Mark Johnson, et~al.
\newblock {SPICE: Semantic Propositional Image Caption Evaluation}.
\newblock In {\em ECCV}, pages 382--398, 2016.

\bibitem{bottom-up}
Peter Anderson, Xiaodong He, et~al.
\newblock {Bottom-Up and Top-Down Attention for Image Captioning and Visual Question Answering}.
\newblock In {\em CVPR}, pages 6077--6086, 2018.

\bibitem{meteor}
Satanjeev Banerjee et~al.
\newblock {{METEOR}: An Automatic Metric for MT Evaluation with Improved Correlation with Human Judgments}.
\newblock In {\em ACL}, pages 65--72, 2005.

\bibitem{cer-etal-2017-semeval}
Daniel Cer, Mona Diab, et~al.
\newblock {SemEval-2017 Task 1: Semantic Textual Similarity Multilingual and Crosslingual Focused Evaluation}.
\newblock In {\em {S}em{E}val}, pages 1--14, 2017.

\bibitem{uniter}
Yen Chen, Linjie Li, Licheng Yu, Ahmed Kholy, Faisal, Zhe Gan, Yu Cheng, et~al.
\newblock {UNITER: Universal Image-text Representation Learning}.
\newblock In {\em ECCV}, pages 104--120, 2020.

\bibitem{flant5}
Hyung Chung, Le Hou, Shayne Longpre, Barret Zoph, Yi Tay, et~al.
\newblock {Scaling Instruction-finetuned Language Models}.
\newblock {\em arXiv preprint arXiv:2210.11416}, 2022.

\bibitem{xlm}
Alexis Conneau, Kartikay Khandelwal, et~al.
\newblock {Unsupervised Cross-lingual Representation Learning at Scale}.
\newblock In {\em ACL}, pages 8440--8451, 2020.

\bibitem{m2trm}
Marcella Cornia, Matteo Stefanini, Lorenzo Baraldi, et~al.
\newblock {Meshed-Memory Transformer for Image Captioning}.
\newblock In {\em CVPR}, pages 10578--10587, 2020.

\bibitem{leic}
Yin Cui, Guandao Yang, Andreas Veit, Xun Huang, and Serge Belongie.
\newblock {Learning to Evaluate Image Captioning}.
\newblock In {\em CVPR}, pages 5804--5812, 2018.

\bibitem{bert}
Jacob Devlin, Ming-Wei Chang, et~al.
\newblock {BERT: Pre-training of Deep Bidirectional Transformers for Language Understanding}.
\newblock In {\em NAACL-HLT}, pages 4171--4186, 2019.

\bibitem{smurf}
Joshua Feinglass and Yezhou Yang.
\newblock {SMURF: SeMantic and linguistic UndeRstanding Fusion for Caption Evaluation via Typicality Analysis}.
\newblock In {\em IJCNLP}, pages 2250--2260, 2021.

\bibitem{capwap}
Adam Fisch, Kenton Lee, Ming Chang, Jonathan Clark, and Regina Barzilay.
\newblock {{C}ap{WAP}: Image Captioning with a Purpose}.
\newblock In {\em EMNLP}, pages 8755--8768, 2020.

\bibitem{simcse}
Tianyu Gao, Xingcheng Yao, and Danqi Chen.
\newblock {SimCSE: Simple Contrastive Learning of Sentence Embeddings}.
\newblock In {\em EMNLP}, pages 6894--6910, 2021.

\bibitem{vild}
Xiuye Gu, Tsung Lin, Weicheng Kuo, and Yin Cui.
\newblock {Open-vocabulary Object Detection via Vision and Language Knowledge Distillation}.
\newblock In {\em ICLR}, 2022.

\bibitem{blind}
Danna Gurari, Yinan Zhao, Meng Zhang, and Nilavra Bhattacharya.
\newblock {Captioning Images Taken by People Who Are Blind}.
\newblock In {\em ECCV}, pages 417--434, 2020.

\bibitem{ort}
Simao Herdade, Armin Kappeler, Kofi Boakye, et~al.
\newblock {Image Captioning: Transforming Objects into Words}.
\newblock In {\em NeurIPS}, volume~32, pages 11137--11147, 2019.

\bibitem{clipscore}
Jack Hessel, Ari Holtzman, et~al.
\newblock {CLIPScore: A Reference-free Evaluation Metric for Image Captioning}.
\newblock In {\em EMNLP}, pages 7514--7528, 2021.

\bibitem{flickr}
Micah Hodosh, Peter Young, and Julia Hockenmaier.
\newblock {Framing Image Description as a Ranking Task: Data, Models and Evaluation Metrics}.
\newblock {\em JAIR}, 47:853--899, 2013.

\bibitem{aoanet}
Lun Huang, Wenmin Wang, Jie Chen, and Xiao Wei.
\newblock {Attention on Attention for Image Captioning}.
\newblock In {\em ICCV}, pages 4634--4643, 2019.

\bibitem{tiger}
Ming Jiang, Qiuyuan Huang, Lei Zhang, Xin Wang, et~al.
\newblock {TIGEr: Text-to-image Grounding For Image Caption Evaluation}.
\newblock In {\em EMNLP}, 2019.

\bibitem{crt}
Motonari Kambara et~al.
\newblock {Case Relation Transformer: A Crossmodal Language Generation Model for Fetching Instructions}.
\newblock {\em IEEE RAL}, 6:8371--8378, 2021.

\bibitem{mid}
Jin Kim et~al.
\newblock {Mutual Information Divergence: A Unified Metric for Multimodal Generative Models}.
\newblock In {\em NeurIPS}, volume~35, pages 35072--35086, 2022.

\bibitem{prmcs}
Yongil Kim, Yerin Hwang, Hyeongu Yun, Seunghyun Yoon, et~al.
\newblock {{PR}-{MCS}: Perturbation Robust Metric for {M}ulti{L}ingual Image Captioning}.
\newblock In {\em EMNLP}, pages 12237--12258, 2023.

\bibitem{wmd}
Matt Kusner, Yu Sun, Nicholas Kolkin, and Kilian Weinberger.
\newblock {From Word Embeddings To Document Distances}.
\newblock {\em PMLR}, 37:957--966, 2015.

\bibitem{vilbertscore}
Hwanhee Lee, Seunghyun Yoon, et~al.
\newblock {{V}i{LBERTS}core: Evaluating Image Caption Using Vision-and-Language {BERT}}.
\newblock In {\em Eval4NLP}, pages 34--39.

\bibitem{umic}
Hwanhee Lee, Seunghyun Yoon, et~al.
\newblock {UMIC: An Unreferenced Metric for Image Captioning via Contrastive Learning}.
\newblock In {\em ACL}, pages 220--226, 2021.

\bibitem{qe_ic}
Tomer Levinboim, Ashish~V. Thapliyal, Piyush Sharma, and Radu Soricut.
\newblock {Quality Estimation for Image Captions Based on Large-scale Human Evaluations}.
\newblock In {\em NAACL}, pages 3157--3166, 2021.

\bibitem{blip}
Junnan Li et~al.
\newblock {BLIP: Bootstrapping Language-image Pre-training for Unified Vision-language Understanding and Generation}.
\newblock In {\em ICML}, pages 12888--12900, 2022.

\bibitem{blip2}
Junnan Li, Dongxu Li, et~al.
\newblock {BLIP-2: Bootstrapping Language-Image Pre-training with Frozen Image Encoders and Large Language Models}.
\newblock In {\em ICML}, 2023.

\bibitem{er-san}
Jingyu Li, Zhendong Mao, Shancheng Fang, et~al.
\newblock {ER-SAN: Enhanced-Adaptive Relation Self-Attention Network for Image Captioning}.
\newblock In {\em IJCAI}, pages 1081--1087, 2022.

\bibitem{rouge}
Chin Lin.
\newblock {{ROUGE}: A Package For Automatic Evaluation Of Summaries}.
\newblock In {\em ACL}, pages 74--81, 2004.

\bibitem{coco}
Tsung Lin, Michael Maire, Serge Belongie, Lubomir Bourdev, Ross Girshick, et~al.
\newblock {Microsoft {COCO}: Common Objects in Context}.
\newblock In {\em ECCV}, pages 740--755, 2014.

\bibitem{roberta}
Yinhan Liu, Myle Ott, Naman Goyal, Jingfei Du, et~al.
\newblock {RoBERTa: A Robustly Optimized BERT Pretraining Approach}.
\newblock {\em arXiv preprint arXiv:1907.11692}, 2019.

\bibitem{dlct}
Yunpeng Luo, Jiayi Ji, Xiaoshuai Sun, Liujuan Cao, et~al.
\newblock {Dual-Level Collaborative Transformer for Image Captioning}.
\newblock In {\em AAAI}, volume~35, pages 2286--2293, 2021.

\bibitem{lens}
Mounica Maddela, Yao Dou, David Heineman, and Wei Xu.
\newblock {LENS}: A learnable evaluation metric for text simplification.
\newblock In {\em ACL}, pages 16383--16408, July 2023.

\bibitem{mabn}
Aly Magassouba, Komei Sugiura, and Hisashi Kawai.
\newblock {Multimodal Attention Branch Network for Perspective-Free Sentence Generation}.
\newblock In {\em CoRL}, pages 76--85, 2019.

\bibitem{marelli-etal-2014-sick}
Marco Marelli, Stefano Menini, Marco Baroni, et~al.
\newblock {A SICK Cure for the Evaluation of Compositional Distributional Semantic Models}.
\newblock In {\em LREC}, pages 216--223, 2014.

\bibitem{ic-survey2}
Yue Ming, Nannan Hu, Chunxiao Fan, Jiangwan Zhou, and Hui Yu.
\newblock {Visuals to Text: A Comprehensive Review on Automatic Image Captioning}.
\newblock {\em JAS}, 9(8):1339--1365, 2022.

\bibitem{cider-r}
Gabriel Oliveira, Esther Colombini, and Sandra Avila.
\newblock {CIDEr-R: Robust Consensus-based Image Description Evaluation}.
\newblock In {\em W-NUT}, pages 351--360, 2021.

\bibitem{bleu}
Kishore Papineni, Salim Roukos, Todd Ward, and Wei Zhu.
\newblock {BLEU: a Method for Automatic Evaluation of Machine Translation}.
\newblock In {\em ACL}, pages 311--318, 2002.

\bibitem{clip}
Alec Radford, Jong~Wook Kim, Chris Hallacy, et~al.
\newblock {Learning Transferable Visual Models from Natural Language Supervision}.
\newblock In {\em ICML}, pages 8748--8763, 2021.

\bibitem{comet}
Ricardo Rei, Craig Stewart, Ana Farinha, and Alon Lavie.
\newblock {COMET: A Neural Framework for MT Evaluation}.
\newblock In {\em EMNLP}, pages 2685--2702, 2020.

\bibitem{att2in}
Steven Rennie, Etienne Marcheret, Youssef Mroueh, Jerret Ross, and Vaibhava Goel.
\newblock {Self-critical Sequence Training for Image Captioning}.
\newblock In {\em CVPR}, pages 7008--7024, 2017.

\bibitem{pac-s}
Sara Sarto et~al.
\newblock {Positive-Augmented Contrastive Learning for Image and Video Captioning Evaluation}.
\newblock In {\em CVPR}, pages 6914--6924, 2023.

\bibitem{bleurt}
Thibault Sellam, Dipanjan Das, and Ankur Parikh.
\newblock {{BLEURT}: Learning Robust Metrics for Text Generation}.
\newblock In {\em ACL}, pages 7881--7892, 2020.

\bibitem{cc3m}
Piyush Sharma et~al.
\newblock {Conceptual Captions: A Cleaned, Hypernymed, Image Alt-text Dataset for Automatic Image Captioning}.
\newblock In {\em ACL}, pages 2556--2565, 2018.

\bibitem{foil}
Ravi Shekhar, Sandro Pezzelle, Yauhen Klimovich, et~al.
\newblock {{FOIL} it! Find One Mismatch Between Image and Language caption}.
\newblock In {\em ACL}, pages 255--265, 2017.

\bibitem{ruse}
Hiroki Shimanaka et~al.
\newblock {RUSE: Regressor Using Sentence Embeddings for Automatic Machine Translation Evaluation}.
\newblock In {\em WMT18}, pages 751--758, 2018.

\bibitem{blind3}
Oleksii Sidorov, Ronghang Hu, Marcus Rohrbach, et~al.
\newblock {TextCaps: A Dataset for Image Captioning with Reading Comprehension}.
\newblock In {\em ECCV}, pages 742--758, 2020.

\bibitem{ic-survey}
Matteo Stefanini, Marcella Cornia, Lorenzo Baraldi, et~al.
\newblock {From Show to Tell: A Survey on Deep Learning-based Image Captioning}.
\newblock {\em PAMI}, 45(1):539--559, 2022.

\bibitem{grit}
Masanori Suganuma, Takayuki Okatani, et~al.
\newblock {GRIT: Faster and Better Image Captioning Transformer Using Dual Visual Features}.
\newblock In {\em ECCV}, pages 167--184, 2022.

\bibitem{transformer}
Ashish Vaswani, Noam Shazeer, Niki Parmar, Jakob Uszkoreit, Llion Jones, et~al.
\newblock {Attention Is All You Need}.
\newblock In {\em NIPS}, volume~30, pages 5998--6008, 2017.

\bibitem{cider}
Ramakrishna Vedantam, Lawrence Zitnick, and Devi Parikh.
\newblock {{CIDEr}: Consensus-based Image Description Evaluation}.
\newblock In {\em CVPR}, pages 4566--4575, 2015.

\bibitem{jaspice}
Yuiga Wada, Kanta Kaneda, et~al.
\newblock {JaSPICE: Automatic Evaluation Metric Using Predicate-Argument Structures for Image Captioning Models}.
\newblock In {\em CoNLL}, 2023.

\bibitem{git}
Jianfeng Wang, Zhengyuan Yang, Xiaowei Hu, Linjie Li, Kevin Lin, et~al.
\newblock {GIT: A Generative Image-to-text Transformer for Vision and Language}.
\newblock {\em TMLR}, 2022.

\bibitem{ofa}
Peng Wang, An Yang, Rui Men, Junyang Lin, Shuai Bai, et~al.
\newblock {OFA: Unifying Architectures, Tasks, and Modalities Through a Simple Sequence-to-sequence Learning Framework}.
\newblock In {\em ICML}, pages 23318--23340, 2022.

\bibitem{open-domain-cqg}
Julia White, Gabriel Poesia, Robert Hawkins, et~al.
\newblock {Open-domain Clarification Question Generation Without Question Examples}.
\newblock In {\em EMNLP}, pages 563--570, 2021.

\bibitem{sat}
Kelvin Xu, Jimmy Ba, Ryan Kiros, et~al.
\newblock {Show, Attend and Tell: Neural Image Caption Generation with Visual Attention}.
\newblock In {\em ICML}, pages 2048--2057, 2015.

\bibitem{sescore2}
Wenda Xu, Xian Qian, Mingxuan Wang, et~al.
\newblock {SESCORE2: Learning Text Generation Evaluation via Synthesizing Realistic Mistakes}.
\newblock In {\em ACL}, pages 5166--5183, 2023.

\bibitem{sescore}
Wenda Xu, Yi-Lin Tuan, Yujie Lu, et~al.
\newblock {Not All Errors are Equal: Learning Text Generation Metrics using Stratified Error Synthesis}.
\newblock In {\em EMNLP}, pages 6559--6574, 2022.

\bibitem{instructscore}
Wenda Xu, Danqing Wang, et~al.
\newblock {INSTRUCTSCORE: Towards Explainable Text Generation Evaluation with Automatic Feedback}.
\newblock In {\em EMNLP}, pages 5967--5994, 2023.

\bibitem{flickr30k}
Peter Young et~al.
\newblock {From Image Descriptions to Visual Denotations: New Similarity Metrics for Semantic Inference over Event Descriptions}.
\newblock {\em TACL}, 2:67--78, 2014.

\bibitem{bartscore}
Weizhe Yuan, Graham Neubig, et~al.
\newblock {BARTScore: Evaluating Generated Text as Text Generation}.
\newblock In {\em NeurIPS}, volume~34, pages 27263--27277, 2021.

\bibitem{vinvl}
Pengchuan Zhang, Xiujun Li, Xiaowei Hu, et~al.
\newblock {VinVL: Revisiting Visual Representations in Vision-language Models}.
\newblock In {\em CVPR}, pages 5579--5588, 2021.

\bibitem{opt}
Susan Zhang, Stephen Roller, Naman Goyal, Mikel Artetxe, et~al.
\newblock {OPT: Open Pre-trained Transformer Language Models}.
\newblock {\em arXiv preprint arXiv:2205.01068}, 2022.

\bibitem{bertscore}
Tianyi Zhang, Varsha Kishore, Felix Wu, Kilian Weinberger, and Yoav Artzi.
\newblock {BERTScore: Evaluating Text Generation with BERT}.
\newblock In {\em ICLR}, 2020.

\bibitem{moverscore}
Wei Zhao et~al.
\newblock {{M}over{S}core: Text Generation Evaluating with Contextualized Embeddings and Earth Mover Distance}.
\newblock In {\em EMNLP-IJCNLP}, pages 563--578, 2019.

\bibitem{region-clip}
Yiwu Zhong, Jianwei Yang, Pengchuan Zhang, Chunyuan Li, et~al.
\newblock {RegionCLIP: Region-based Language-image Pretraining}.
\newblock In {\em CVPR}, pages 16793--16803, 2022.

\end{thebibliography}
}

\clearpage
\appendix
\setcounter{page}{1}
\maketitlesupplementary

\section{Additional Related Work}

\paragraph{Image captioning}
Numerous studies have been conducted in the domain of image captioning \cite{sat,ort,m2trm,dlct,er-san}.
For instance, GRIT \cite{grit} adeptly leverages both grid and region-based features for enhanced image captioning, thereby eliminating the need for conventional CNN-based detectors.  
BLIP-2 \cite{blip2} is a novel pre-training approach that efficiently uses LLMs that outperforms models including Flamingo \cite{flamingo} with notably fewer trainable parameters.
This field includes applications in various domains, such as aiding persons with vision impairment \cite{blind, blind2, blind3} and robotics \cite{mabn,crt}. 
Survey papers such as \cite{ic-survey, ic-survey2} offer a comprehensive overview of image caption generation, including models, standard datasets, and evaluation metrics. Specifically, they provide a comprehensive summary of various automatic evaluation metrics, including similarity-based and learning-based metrics \cite{wmd, moverscore, bertscore}.

\vspace{-2.5mm}
\section{Polaris Dataset}
\label{appendix:stat}

\subsection{Meta-Analysis}
\label{appendix:dataset-meta}

\begin{figure}[t]
    \centering
    \includegraphics[width=\linewidth]{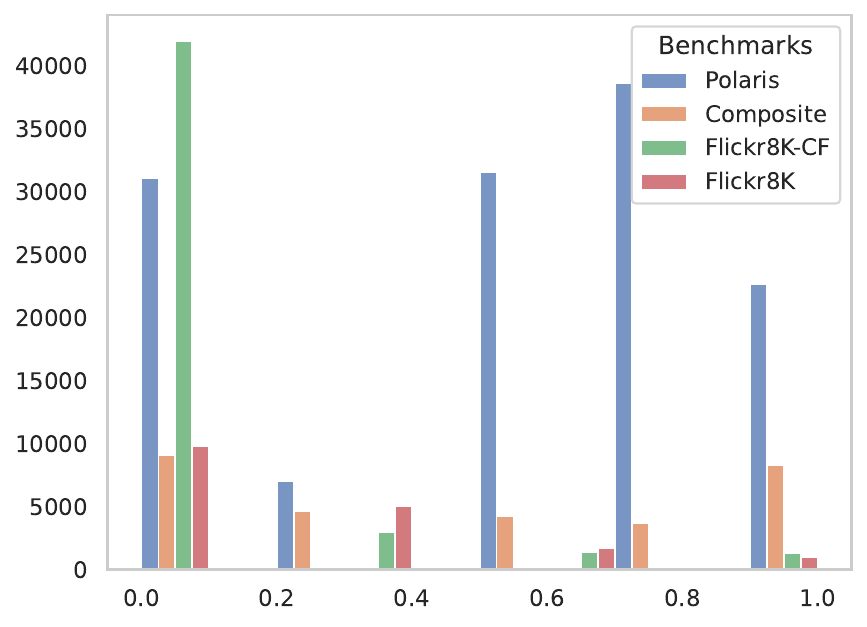}
    \caption{Score distributions of human judgments in \textbf{\textcolor{composite}{Composite}}, \textbf{\textcolor{flickr8k-ex}{Flickr8K-Expert}}, \textbf{\textcolor{flickr8k-cf}{Flickr8K-CF}}, and our \textbf{\textcolor{polaris}{Polaris}} dataset. All scores were normalized from 0 to 1. Polaris distinguishes itself from other datasets by encompassing a vast collection of captions and integrating a broad spectrum of human judgments.}
    \label{fig:hist}
    \vspace{-5mm}
\end{figure}

As pointed out in \cite{umic}, there are issues with utilizing existing datasets such as Flickr8K and Composite for training purposes.
Fig.\ref{fig:hist} shows the score distributions of human judgments in Composite, Flickr8K-Expert, Flickr8K-CF, and our proposed Polaris dataset.
For the Flickr8K dataset, the majority of scores fall below 0.4, as the candidate captions were sourced from a reference caption pool through an image retrieval system.
Moreover, the Flickr8K datasets do not contain captions generated by models, which presents an issue from the perspective of the domain mismatch because our aim is to build an automatic metric for image captioning. Consequently, we argue that Flickr8K-Expert and Flickr8K-CF are not suitable for training metrics.
Furthermore, the human judgments in Flickr8K-CF were provided using a binary scheme, that only allowed responses categorized as ``yes'' or ``no.''
This method is problematic because of its lack of granularity and its propensity to force evaluators into making overly simplistic judgments. For instance, a disparity exists in quality for captions that describe content and our method may not be able to adequately evaluate captions of varying quality.

In the case of the Composite dataset, we note its exceptionally few human judgments. This paucity of data renders it inadequate for developing a practical metric. Moreover, as pointed out in \cite{umic}, each sample's score was determined by a single annotator, leading to potentially biased outputs. Upon manual examination of the captions, \cite{umic} also pointed out that these captions are often coarsely generated.

\subsection{Statistics and Details}
The Polaris dataset includes 13,691 images accompanied by 131,020 generated captions. Additionally, it contains 262,040 references. All sentences are in English. To minimize biases in evaluations and achieve more balanced judgments compared with other datasets, we engaged multiple human evaluators to evaluate each caption. Specifically, each generated caption was evaluated by approximately eight different evaluators. 
The generated captions encompass a vocabulary of 3,154 unique words, with a total of 1,177,512 words. On average, each caption is composed of 8.99 words. By contrast, the reference captions have a vocabulary of 22,275 unique words, with a word count of 8,309,300. Each reference caption, on average, consists of 10.7 words.

In the Polaris dataset, the training, validation, and test sets consist of 78,631, 26,269, and 26,123 samples, respectively. 
We used the training set to train the model, the validation set for hyperparameter tuning, and the test set to evaluate the performance of the model.

\subsection{Image Captioning Models}

The Polaris dataset comprises captions generated by the following 10 standard models. Table \ref{tab:ic-models} summarizes these image captioning models. We selected these models as they are standard image captioning models. Additionally, we also chose older models to ensure diversity in the quality of their output sentences. 

\begin{table}[H]
    \centering
    \normalsize
    \begin{tabular}{lc}
    \hline
{Year} & {Venue} \\ \hline
{$\mathrm{BLIP}$-$\mathrm{2_\mathrm{flan}}$ \cite{blip2}} & {ICML'23} \\ 
{$\mathrm{BLIP}$-$\mathrm{2_\mathrm{opt}}$ \cite{blip2}} & {ICML'23} \\ 
{$\mathrm{GRIT}$ \cite{grit}}  & {ECCV'22} \\
{$\mathrm{OFA}$ \cite{ofa}}  & {ICML'22} \\
{$\mathrm{GIT}$ \cite{git}}  & {TMLR'22} \\
{$\mathrm{BLIP_\mathrm{large}}$ \cite{blip}} & {ICML'22} \\
{$\mathrm{BLIP_\mathrm{base}}$ \cite{blip}} & {ICML'22} \\
{$\mathrm{VinVL}$ \cite{vinvl}} & {CVPR'21} \\
{$\mathrm{\mathcal{M}^2}$-$\mathrm{Transformer}$ \cite{m2trm}} & {CVPR'20} \\
{$\mathrm{SAT}$\cite{sat}} & {CVPR'15} \\ \hline
    \end{tabular}
    \caption{Image captioning models used for the Polaris dataset.}
    \label{tab:ic-models}
    \vspace{-3mm}
\end{table}

\section{Error Analysis}

To investigate the limitations of the proposed metric, we analyzed 100 instances where the method did not perform as expected.
We define failed cases as samples that satisfy the condition $|y - \hat{y}| \geq \theta$. In this study, we held $\theta$ at a value of 0.25, corresponding to the step size when normalizing a five-level evaluation.
Table \ref{tab:error} categorizes the failed cases. The causes of failure can be grouped into seven categories:
\begin{enumerate}
\vspace{-2mm}
\setlength{\parskip}{0cm}
\setlength{\itemsep}{0cm}

\item[(i) ] Overestimation of captions lacking details (OCLD):
This category pertains to instances where the proposed metric assigned higher scores to captions that lacked vital details, missing critical aspects of the images.

\item[(ii) ] Overestimation of captions with incorrect details (OCID):
This category refers to instances where the proposed metric inaccurately assigned higher scores to captions containing incorrect or misleading details.

\item[(iii) ] Underestimation of captions where the focus areas differ from the references (UCFA):
This category refers to instances where the proposed metric assigned lower scores to captions that, although accurate, focused on areas different from the references.

\item[(iv) ] Serious Errors (SE):
This category encompasses instances where the evaluation deviated greatly from human judgments, being much higher or lower.

\item[(v)] Overestimation of captions with grammatical inaccuracies (OSGI):
This category refers to instances where the proposed metric erroneously assigned higher scores to captions that, while potentially accurate in content, contained grammatical errors.

\item[(vi)] Annotation errors (AE):
This category pertains to instances where human evaluations proved to be inaccurate. These evaluations were either higher or lower than what could be reasonably expected.

\item[(vii)] Others:
This category encompasses miscellaneous errors that do not fit into the aforementioned categories.

\end{enumerate}

From Table \ref{tab:error}, it can be inferred that the main bottleneck was the overestimation of captions that lacks detail. 
As mentioned in Section \ref{sec:discussions}, a possible solution could be to enhance the fine-grained alignment \cite{region-clip}.

\begin{table}[t]
    \centering
    \setlength{\tabcolsep}{2pt}
    \normalsize
    \begin{tabular}{cp{6cm}c}
    \hline
{Errors} & {Description} & {\#Error} \\ \hline
{OCLD} & {Overestimation of captions lacking details} & {29} \\
{OCID} & {Overestimation of captions with incorrect details} & {22} \\
{UCFA} & {Underestimation of captions where the focus area differs from
the reference} & {15} \\
{SE} & {Serious errors (e.g., assigning a higher score to captions with
major mistakes)} & {11} \\
{OGI} & {Overestimation of captions with grammatical inaccuracies} &
{11} \\
{AE} & {Annotation errors in human judgments} & {9} \\
{Others} & {Miscellaneous or less common errors} & {3} \\ \hline
{Total} & {---} & {100} \\ \hline
    \end{tabular}
    \caption{Categorization of failed samples.}
    \label{tab:error}
    \vspace{-5mm}
\end{table}

\section{Implementation Details}
\label{appendix:impl}

Table \ref{tab:failure} shows the experimental settings for the proposed method.
We trained our model on a Tesla A100 GPU, and the training time was approximately 4.6 hours.
To measure the inference time, we tested our metric on a system equipped with a GeForce RTX 3090 and an Intel Core i9-10900KF. The inference times per sample for SPICE \cite{spice}, RefPAC-S \cite{pac-s}, and Polos were $16.6$ ms, $4.45$ ms, and $6.91$ ms, respectively.  Notably, Polos operates at a speed $2.4$ times faster than SPICE.
Furthermore, in the 6.91 ms processing time of Polos, CLIP took 4.90 ms, RoBERTa 1.96 ms, and MLP just 0.054 ms.
We used early stopping in our model to optimize for the highest Kendall's $\tau$. At each epoch, we evaluated Kendall's $\tau$ on the validation set. If no improvement was observed over five consecutive epochs, we stopped training. Subsequently, we evaluated the model's performance using the test set, referring to the epoch where the validation set achieved its best $\tau$ value.

\begin{table}[H]
    \centering
    \caption{Experimental settings for the proposed method.}
    \normalsize
    \begin{tabular}{cc}
    \hline
        {Batch size} & {64} \\
        {Optimizer} & {Adam ($\beta_1 = 0.9, \beta_2 = 0.98)$} \\
        {Learning Rate} & {$3.0 \times 10^{-5}$} \\ \hline
    \end{tabular}
    \label{tab:failure}
\end{table}

\section{Additional Qualitative Results}
\label{appendix:qualitive}


Fig.\ref{fig:example1} and Fig.\ref{fig:example2} provide additional comparisons between the CLIPScore family and Polos. We observed that the CLIPScore family tends to overestimate scores. Specifically, RefCLIP-S and RefPAC-S may not effectively compare references and a candidate. Although CLIP-S could not show overestimation, this does not imply adequacy in caption evaluation. 
Rather, it may signal estimation deficiencies, especially for longer captions, stemming from poor alignment between words and images, as CLIP relies heavily on the alignment between image and language features.

\input{pages/additional_examples}

\end{document}